\documentclass[conference]{IEEEtran}
\IEEEoverridecommandlockouts
\usepackage{cite}
\usepackage{amsmath,amssymb,amsfonts}
\usepackage{algorithmic}
\usepackage{graphicx}
\usepackage{textcomp}
\usepackage{xcolor}
\usepackage[hidelinks]{hyperref}
\usepackage[utf8]{inputenc}
\def\BibTeX{{\rm B\kern-.05em{\sc i\kern-.025em b}\kern-.08em
    T\kern-.1667em\lower.7ex\hbox{E}\kern-.125emX}}
\begin{document}

\title{Robot Soccer Kit:
Omniwheel Tracked Soccer Robots for Education
}

\author{\IEEEauthorblockN{Grégoire Passault}
\IEEEauthorblockA{\textit{LaBRI - DART - Rhoban} \\
\textit{University of Bordeaux}\\
Bordeaux, France \\
gregoire.passault@u-bordeaux.fr}
\and
\IEEEauthorblockN{Clément Gaspard}
\IEEEauthorblockA{\textit{LaBRI - DART - Rhoban} \\
\textit{University of Bordeaux}\\
Bordeaux, France \\
clement.gaspard@u-bordeaux.fr}
\and
\IEEEauthorblockN{Olivier Ly}
\IEEEauthorblockA{\textit{LaBRI - DART - Rhoban} \\
\textit{University of Bordeaux}\\
Bordeaux, France \\
olivier.ly@u-bordeaux.fr}
}

\IEEEoverridecommandlockouts \IEEEpubid{\makebox[\columnwidth]{978-1-6654-8217-2/22/\$31.00~\copyright2022 European Union \hfill} \hspace{\columnsep}\makebox[\columnwidth]{ }}

\maketitle

\IEEEpubidadjcol

\begin{abstract}
    Recent developments of low cost off-the-shelf programmable components, their modularity, and also 
    rapid prototyping made educational robotics flourish, as it is accessible in most schools today. 
    They allow to illustrate and embody theoretical problems in practical and tangible applications, and
    gather multidisciplinary skills.  They also give a rich natural context for 
    project-oriented pedagogy. However, most current robot kits all are limited to  
    egocentric aspect of the robots perception. This makes it difficult to access more 
    high-level problems involving {\em e.g.} coordinates or navigation. In this paper we introduce an educational 
    holonomous robot kit that comes with an external tracking system, which lightens the constraint on 
    embedded systems, but allows in the same time to discover high-level aspects of robotics, otherwise unreachable.
\end{abstract}


\section{Introduction}

Educational robotics is a field promoting the use of robots as tools to engage learners on practical
applications, problems, and sometime competitions. This approach can be backed up by constructionist and
experimental learning theories.
A lot of educational robotics platforms recently emerged and are now used in classrooms.

Most of educational robotics kit provide platforms including actuators and sensors along with possibilities to
program them using adapted interfaces \cite{evripidou2020educational}.
No-code or visual programming languages (VPL) received a lot of interest, since it allows a fast onboarding
in the basics of programming. We can mention Scratch \cite{resnick2009scratch} and Google's Blockly that offers
well known puzzle-like blocks environments.


Some other projects like Arduino emphasizes the "DIY empowerment", where users are given some actuators
and sensors, and the ability to integrate them without extensive knowledge about low-level aspects.

However, almost all those robots only use egocentric sensors (distance sensors, camera, IMU, lidar etc.)
giving partial information on the robot localization in the world.
Finding the robot's localization can be achieved using extensive filtering like Kalman, particle filters,
factor graph or SLAM. But, all these methods have theoretical prerequisites (some of them being ongoing 
research) and are impossible to address with young students.

Even if manipulating such sensors is interesting as makes students discover the world of embedded systems, 
most applications are limited to what can be done without a robot's localization, involving for instance 
dead reckoning or line/color following. With such technologies, students have no access to topics involving
coordinates and navigation  like path planning, obstacles avoidance or multi-robots strategies. Yet these last
ones would constitute rich subjects to apply their mathematical backgrounds.

\begin{figure}[htb]
    \begin{center}
        \includegraphics[width=0.8\linewidth]{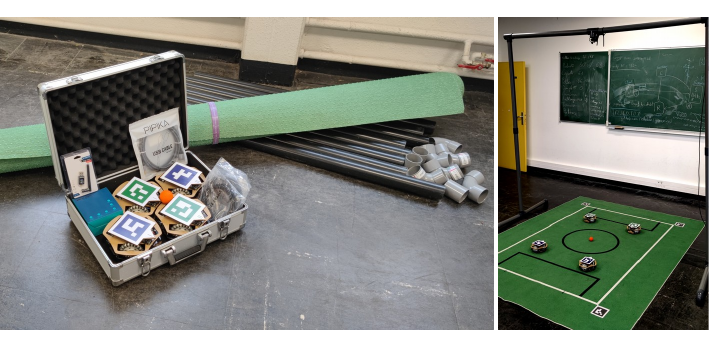}
        \caption{\label{kit}
            A packed and assembled overview of the kit.
        }
    \end{center}
\end{figure}


Our initiative roots in RoboCup \cite{kitano1997robocup}, the biggest international robotics competition, where teams meet to compete in various leagues, including soccer games, rescue or service robotics.
One of the founding league is the soccer competition, intended to ultimately develop a team of robots able to defeat
the best teams of humans.
Initially intended for researchers, the competition expanded to young students with RoboCup Junior \cite{sklar2002robocupjunior}, which is now a very well established world-wide educational project. 
Major leagues for researchers aims at unlocking scientific difficulties like robust indoor navigation or biped walking, while junior leagues provide today an unrivaled international framework (technically and socially) for students to learn about how to design and program autonomous robots.

The RoboCup Junior also features a soccer league, where two teams of two wheeled robots face each other
on a small field. Participating in such a league requires to first design and build the robots, which has
a huge cost for newcomers that can be impossible to tackle in educational environment with no mechatronics background.
One major league, the RoboCup SSL \cite{weitzenfeld2014robocup},
took a different stance on this soccer challenge, by using robots localized by cameras and controlled by external
(non-embedded) computers, with an overall access to detection information.

\begin{figure}[htb]
    \begin{center}
        \includegraphics[width=0.3\linewidth]{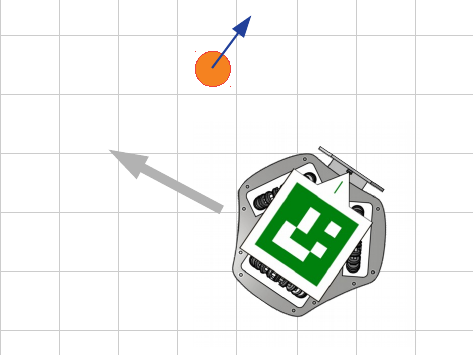}
        \caption{\label{maneuver} 
            To kick the ball in the direction given by the arrow, an omnidirectional robot can perform an
            immediate translation, where its two-wheeled counterpart would require a complex maneuver.
        }
    \end{center}
\end{figure}

Inpired by these concepts, we introduce Robot Soccer Kit\footnote{see
\url{https://robot-soccer-kit.github.io/} for all material,
including videos} (RSK): a novel pedagogical kit (Figure \ref{kit})
that includes holonomous robots, together with an external
vision-based tracking system.
The tracking system gives access to coordinates to the students, and
opens the world of navigation to him.
Also, we choose holonomous robots because their motion is a direct
application of classical mathematical/kinematics concepts (rotation
angle and speed vector). Classical two-wheeled robots are easy to
build and understand, but come with so-called nonholonomic constraints
since the chassis can’t move sideways (see \cite{lynch2017modern},
Chapter 13). Such robots can reach any configuration, but may require
complex maneuver. On the other hand, omnidirectional robots are harder
to build and understand, but easier to program (see Figure
\ref{maneuver}).
Let us note that the system is not intrinsically
associated to soccer, we will see that it can enforce other applications,
by designing fora field representing for {\em e.g} a logistic warehouse or a restaurant
with clients to serve without changing anything else.

We first present the global architecture along with pedagogical
arguments for the choices that were made, then present possible pedagogical scenarios based on this kit, and
finally discuss about the possibility of organizing soccer competitions games with this setup (one could think
of this as a "pocket RoboCup SSL").


\section{Architecture}

\subsection{Overview}

\begin{figure}[htb]
    \begin{center}
        \includegraphics[width=0.7\linewidth]{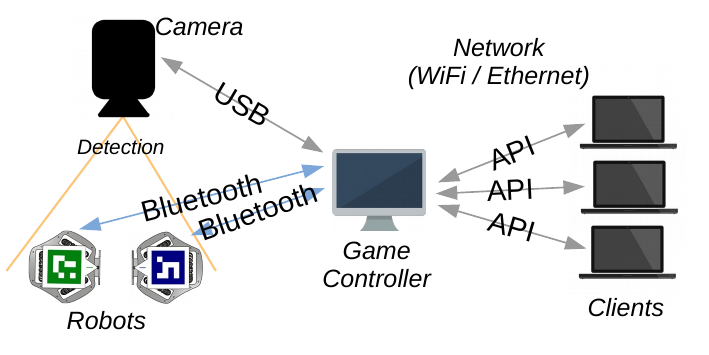}
        \caption{\label{architecture}
            Overview of the architecture
        }
    \end{center}
\end{figure}

Here is the architecture we propose:

\begin{itemize}
    \item{
        Omnidirectional robots evolving on a field,
    }
    \item{
        One wide-angle camera tracking everything from above,
    }
    \item{
        A software (the \textit{Game Controller}) communicating with both camera and the robots,
        and exposing an high-level API,
    }
    \item{
        Clients connecting to the API (optionally through some IP network).
    }
\end{itemize}

This approach depicted on figure \ref{architecture} is very centralized, but once the \textit{Game Controller} is
setup, only a very lightweight API client is required to access detection information and control the robots.

The footprint of the setup is 2.5x2m on the ground, with an height of 2m for the camera. The total price tag is
about 1250 USD.

In a class configuration, this can be convenient to lightens software setup, since only a small software package is required on most of the computers.

In a competition configuration, this means that the teams only need to be on the same network as the
\textit{Game Controller} and run their program to control the robots. With this approach, every communication with
the camera or robot is effectively achieved by the \textit{Game Controller}, which means for example that it is
always possible to preempt the robots from being controlled by the users (for a referee action for instance).

\subsection{Robots design}

\begin{figure}[htb]
    \begin{center}
        \includegraphics[width=0.35\linewidth]{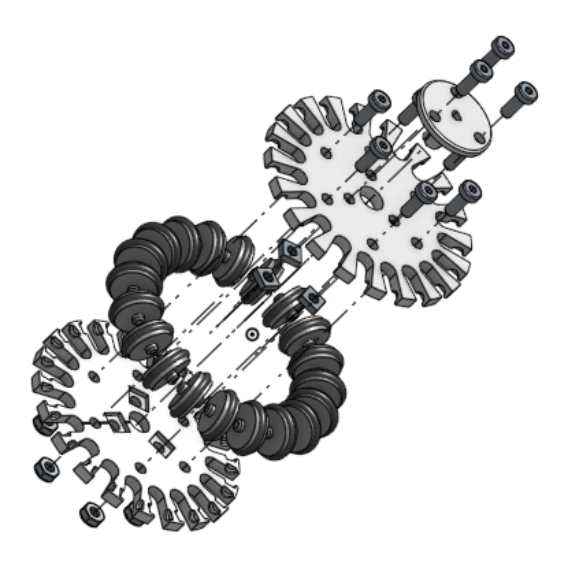}
        \caption{\label{exploded} 
            An exploded view of the custom wheels
        }
    \end{center}
\end{figure}

Robots in RSK are equiped with custom omnidirectional wheels, each one using 20 bearings and 3D printed
parts that can be made using low-cost filament printers (see figure \ref{exploded}). The inverse kinematics is directly
implemented in the robots firmware which allows them to be controlled directly with chassis target speed (instead
of wheels target speeds). Goal wheel speeds can be reached by servoing with to incremental sensors.

At maximum speed, robots can translate at 20 cm/s and rotate at 180 °/s.
Communication with the robots is performed using Bluetooth. Most computers are natively equipped of
Bluetooth devices, and the others can use an external USB dongle.


\subsection{Kickers}

In soccer games, the ball is moving faster than players, which makes the player's strategy non trivial and
enforce the development of some anticipations. RSK robots are equipped with kickers devices,
for this purpose, some capacitors are charged with a boosted voltage (here, 20V in 4000µF), and discharged
in a solenoid plunger. It has been designed for not being dangerous for users.

The duration of the impulse in the plunger can be adjusted to change the power of the kick (it is
currently capped at 5ms).

We use polyurethane orange golf balls, which weights 8g and 42 mm diameter. With such ball and field, robots are
able to kick the ball at 80\% of the field length, with an initial speed of approximately 1 m/s (it can still
be tracked at this speed as experimented on figure \ref{ballspeed}).

\subsection{Vision-based tracking}

\begin{figure}[htb]
    \begin{center}
        \includegraphics[width=0.7\linewidth]{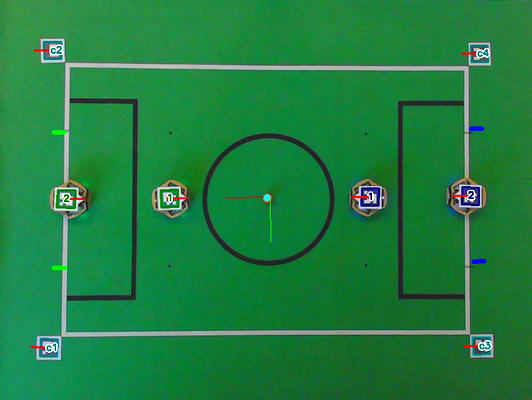}
        \caption{\label{field}
            The result of detection system (processed at 30Hz) with live annotations.
        }
    \end{center}
\end{figure}

External tracking based on vision is both cheap and easy to setup. To
even simplify the tracking of objects, we use ArUco \cite{romero2018speeded} fiducial markers, which are
implemented in \cite{opencv_library}.
Those markers are square images containing data optimized so that its position and orientation is detected.
We use one per robot and a static one at each corner of the field (see figure \ref{field}).

Knowing the dimension and position of the field corner fiducial markers, we then have up to 16 points visible
with a ground-truth position. We use those points to find the best-match homography to transform coordinates
from pixels on the camera to the field coordinate system. Since finding an homography requires a minimum of only
4 points \cite{dubrofsky2009homography} and we use 16, we can check that the calibration is successful by
comparing the expected position of known points with their detected counterparts.

We also check that the whole field is visible by the camera by testing that its corners are projected inside
the image using the inverse of the homography.

The ball is detected using its orange hue with coarse thresholds in the HSV color space. Nothing else was kept
anywhere close to this hue on purpose to simplify calibration.

The camera is positioned perpendicularly to the ground, in order to ensure uniform resolution for objects
on the field.

Note that the graphics of the field is totally aesthetic and can be replaced by any other pattern as long as the
fiducial markers are kept in the corners with the same dimensions, which allows to consider any other contexts ({\em e.g.} a warehouse).

Also, extra markers can be used to tag obstacles that can be put on the ground.
Since the geometry of those obstacles and their pose are both known, they can be used to implement algorithms
involving navigation or path finding.

\begin{figure}[htb]
    \begin{center}
        \includegraphics[width=0.6\linewidth]{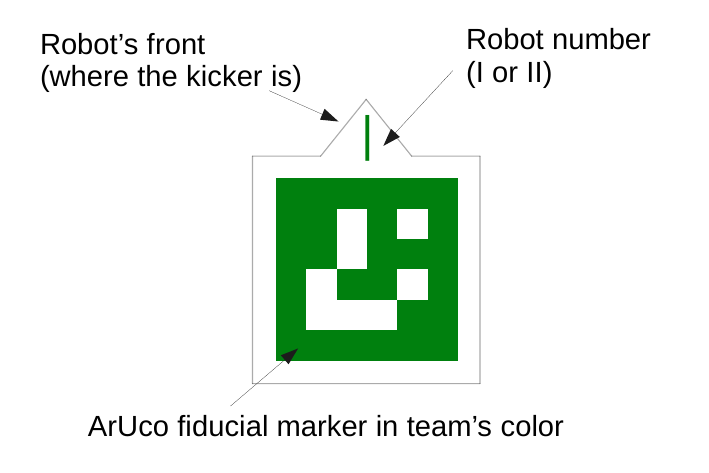}
        \caption{\label{robotmarker}
            Design of robot's markers. The marker's color, front arrow and robot number are
            additional informations for human spectators.
        }
    \end{center}
\end{figure}

The robots' markers includes extra informations for humans spectators to understand what robots (color, number and
orientation) are on the field (see figure \ref{robotmarker}).

On a \textit{Intel(R) Core(TM) i7-7820HQ CPU @ 2.90Ghz}, the system can process the whole detection at 30 frames
per seconds, acquiring images of 1280x720 cropped to 820x635. At this scale, each pixel on the screen represents
3mm.

\subsection{Development environment}

Computer Science is taking an increasingly important place in education.
In France, two years ago, a new material\footnote{NSI: \textit{Numérique et Sciences Informatiques} / Digital and Computer Science} 
dedicated to that purpose  has been introduced in high school.
Its curricula \footnote{\url{https://www.education.gouv.fr/bo/19/Special8/MENE1921247A.htm}} is mentioning 
the fact that ``a language that is easy to use, interpreted, concise, open and free, multi-platform, widely used, 
rich  in adapted libraries and benefiting from a vast community of authors in the educational world is to be preferred''.

We chose Python, which is also mentioned explicitly, as it meets all these criteria and is widely used around
the world in educational or professional projects.
Moreover, according to the fact that Python is and will be more and more used in education, the development of a 
Python library guarantee compatibility and ease of installation with the majority of computer 
equipment in the education sector.

As explained before, RSK is structured around a \textit{Game Controller}, communicating with robots and
exposing a remote API that Python clients can use to:

\begin{enumerate}
    \item{
        {\bf Obtain all detection informations}:
        robots positions and orientations, ball position 
        (expressed in meters and radians in the field frame)
    }
    \item{
        {\bf Send chassis speed target}:
        the client can give $(\dot x, \dot y, \dot \theta)$ target for the robots it is allowed to control
    }
    \item{
        {\bf Send kick orders}:
        the client can ask a robot to perform a kick with a given impulse duration
    }
\end{enumerate}

On the client side, we also provide a {\bf goto} method, that can be use to send a robot to
the $(x, y, \theta)$ target position. This pre-implemented method only uses above primitives to
compute an error and perform servoing to bring the robot to the desired absolute position.

\section{Pedagogical sequences} \label{pedasequ}

In this section, we introduce examples of pedagogical sequences, that are objective-oriented activities
that can be performed in-class with students.

The goal is to build step-by-step exercises eventually leading to having two soccer playing robots.
As much as possible, we try to get the students facing the shortcomings of their solution by practice.
In most cases, we can alternate between working on robots and working on paper and whiteboard.

The following examples are meant to be multidisciplinary. Indeed, they call on different 
notions studied in Engineering Sciences, Physics but also in Computer Science. They also are
multi-level with different stage of difficulties. The majority of them are calibrated for teaching in high school, 
however, some can also be made slightly more complex for being used in higher education.

\subsection{Given that the robot is at the center of the field, make it face the ball, using the "goto" primitive}

This introductory exercise is a way to reminds how cartesian and polar coordinates works. More
precisely, it is implicitly about converting cartesian coordinates $(x, y)$ to the $\theta$
polar angle. Students with trigonometry background might typically come up with the solution
$tan^{-1} (\frac{y}{x})$.

After implementing it and trying on the robot, they might observe that it doesn't always work.
Indeed, it only works in quadrants I and IV, and it is not robust to division by zero.

A robust solution can then be elaborated by testing all the cases manually, and the function
$atan2$ can eventually be mentioned. The latter is an essential function to know when programming
geometry and is a prerequisite for almost other exercises we can think of.

\subsection{Given that the robot is at the center of the field, make it face the ball using rotational speed primitives}

Previous exercise example is a prerequisite for that one, where we propose to use direct
chassis target speed instead of the pre-implemented "goto" provided primitive.

First, the students can be steered to a solution involving an error
$\delta_\theta = \theta_{target} - \theta$ and try fixing it using three cases
(turn left, stop moving, turn right). Then, the intuition that the bigger the error, the
bigger the speed we want to inject to correct it can be used to introduce proportional
servoing.

All this can be tested on the robot. With previous implementation, if $\theta = -3 rad$
and $\theta_{target} = 3 rad$, then $\delta_\theta = 6 rad$, which causes the robot to
correct the error the other way around we would except. This illustrate the typical
problem of \textit{angle wrapping} (the angles need to be wrapped between $-\pi$ and
$\pi$ to be used as an orientation error).

\subsection{Get the robot going toward the ball facing it, and kicking it}

Again, this is incremental question easier to address after the previous exercise has been solved. An approach can
be to use the orientation servoing to face the ball, except that since the robot is moving it should take in account
that the target orientation should be expressed around the robot origin instead of the field origin.

Going forward can be achieved using some conditions on the angular error (if the error is reasonably small, go
forward, else, simply rotate), but rotation can be controlled all the time to keep the ball aligned with the front
of the robot. Deciding when to kick can be done using thresholds on distance from robot to ball.


\subsection{Wherever the ball and the robots initially are, move the robot and make it kick so that it scores a goal in
opponent goals using the "goto" primitive (implement an attacking robot)}

This exercise is not difficult in the sense that the students' toolbox contains all the
theoretical tools to solve it. However, they need to assemble the knowledge they have and
turn it into an algorithm.

\begin{figure}[htb]
    \begin{center}
        \includegraphics[width=0.8\linewidth]{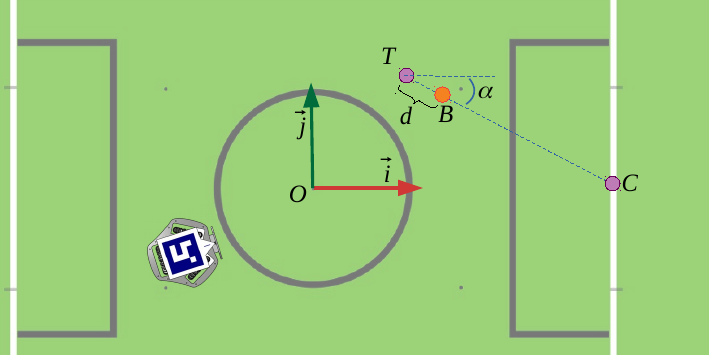}
        \caption{\label{problem}
            An example of pedagogical problem: compute $T$ and $\alpha$ to bring the robot at a distance
            $d$ of the ball, facing the center $C$ of opponent goals.
        }
    \end{center}
\end{figure}

We can represent the problem with figure \ref{problem}. The goal is to derive the target
position $T$ and orientation $\alpha$ for the robot from the other variables of the problem.
The center of goals (point $C$) is a known constant and the position of the ball (point $B$)
is given by the detection system. For the robot not to push the ball when approaching it,
we need to take it away from the ball with a constant distance $d$.


This can then be implemented and tested on the real robot. Triggering the kick requires for the
robot to go forward for a small time step to ensure the ball is touching the kicker. This
mechanism, along with the value of $d$, can be tuned on the real setup to optimize performances.

\subsection{Knowing the position and orientation of opponent robot, place a goal so that it stops the ball (implement a goalee)}

One way to implement this is to use a symmetrical approach to the attacking robot, using again figure \ref{problem}.

The first step to consider is that we need an estimate of which opponent robot is the most likely to attack us and
try to score a goal. Then, one can use its orientation to extrapolate where the ball would arrive if it was kicked
by this robot, in order to place the goalee efficiently.

We could propose to students to search for an affine function $f(x) = ax + b$ that pass through $B$ and has a
slope matching the orientation of attacking robot, and then check the value of $f(\frac{l}{2})$, the function
intersection with the half-field length.

In implementation pitfalls, the opponent orientation that are not in $[-\frac{\pi}{2}, \frac{\pi}{2}]$ might
be ignored, and the goalee should be capped inside its goals where it is more relevant to stay.

\subsection{Re-implement the "goto" primitive}

The "goto" primitive mentioned before allows a fast onboarding. Target speeds that are sent to the
robots are intrinsic chassis speed expressed in the robot frame, thus, sending the robot to a given target
requires changing the frame in which this target is expressed.

This is a perfect and immediate application of the concept of frames, that is a core skill in robotics.

\subsection{Create a model to estimate the future position of the ball when it comes to a stop}

When the robots the strongest kick, it takes about 3 seconds for the ball to come to a stop. If a robot is
moving toward the ball without any anticipation, it will waste most of its time, sometime moving
forward then backward because of the motion of its target.

\begin{figure}[htb]
    \begin{center}
        \includegraphics[width=1.\linewidth]{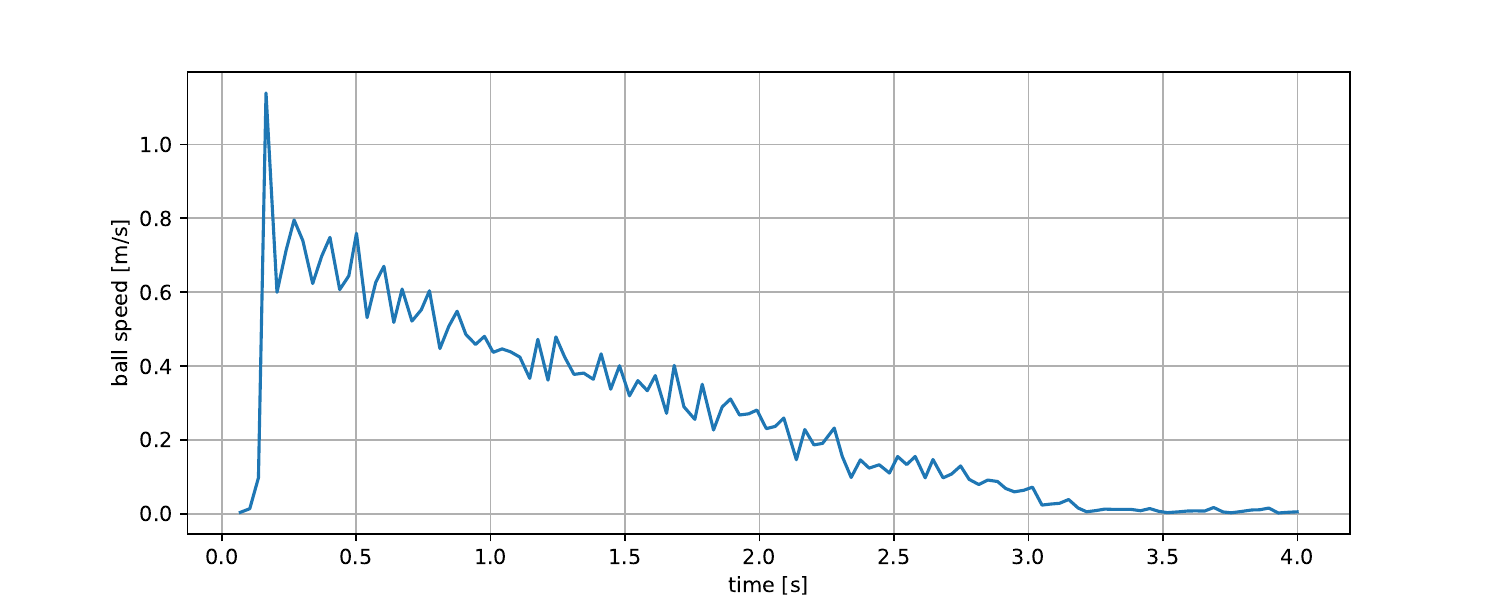}
        \caption{\label{ballspeed}
            A log of estimated ball speed using finite differences after a maximum
            kick. We can observe that the ball deceleration can be reasonably
            approximated by a linear constant of $0.25 m.s^{-2}$.
        }
    \end{center}
\end{figure}

Students can be asked to try to build a predictive model to try to estimate the future position of the ball
(when it comes to a stop) to account for that and give a better target to a moving robot.
They can then be introduced to data logging and plotting, which all can be done very conveniently since
all the aspects of programming are performed remotely through the API.
Using finite difference, they can then plot the estimated speed of the ball versus time and obtain a
figure similar to figure \ref{ballspeed}.

Thanks to this, we can notice that using a constant deceleration $\ddot x$ is a good approximation of the ball
dynamics. By estimating the position $x$ and the speed of the ball $\dot x$ along its axis, the
times when it comes to a stop $t_{stop} = \frac{\dot x}{\ddot x}$ can be estimated, and plugged in the
integration $x + \dot x t_{stop} + \frac{1}{2} \ddot x t_{stop}^2$ to estimate the position of the ball when
it will comes to a stop.

\subsection{By logging data, adjust the duration of solenoid impulse to control the kicked speed of the ball instead}

The relation between the duration of the impulse in the solenoid and the speed of the ball once kicked is both
noisy and not linear at all.

\begin{figure}[htb]
    \begin{center}
        \includegraphics[width=1.\linewidth]{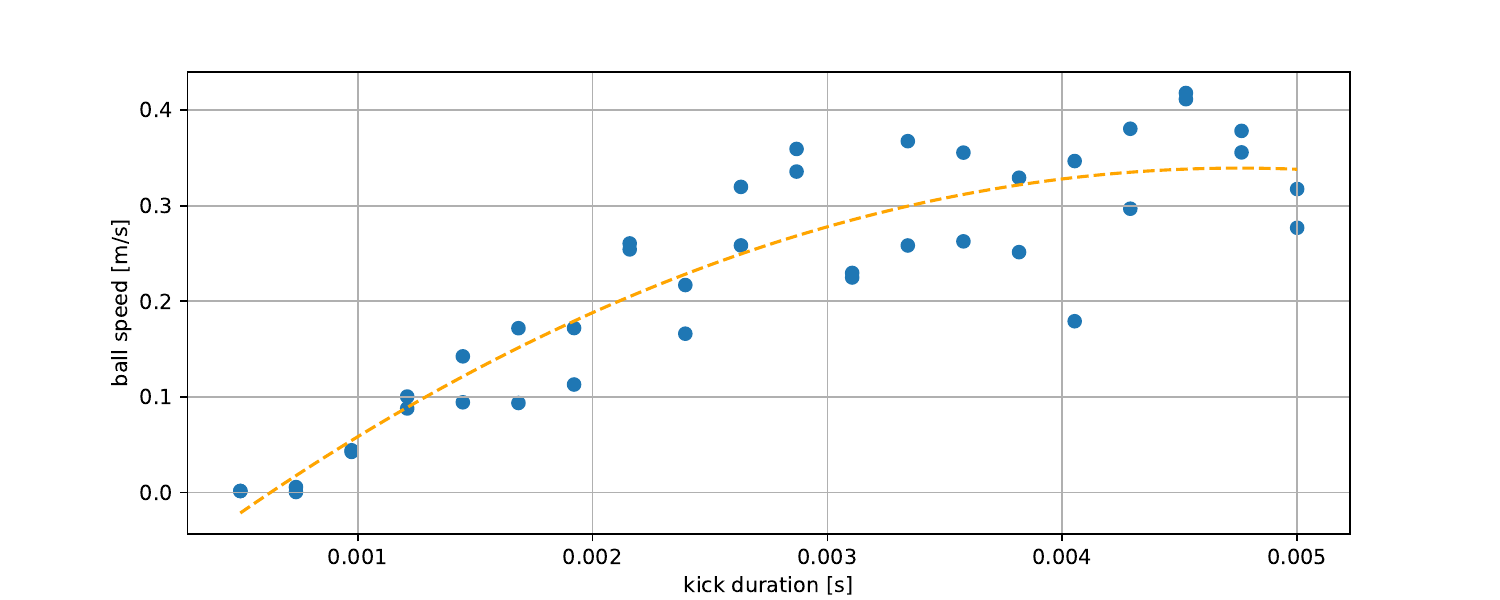}
        \caption{\label{kickspeed}
            Estimated ball speed versus kick impulse duration in the solenoid. A model
            can be fitted to reverse this function.
            However, the measure is very sensitive to initial condition, resulting in
            a lot of sampling noise. The dashed orange line is the best-fit for a second
            order polynomial model.
        }
    \end{center}
\end{figure}

To have a better control over the kicks performed by robots, it is a good idea to calibrate the kicker by sampling
several kicks and try to fit a model over the logged data. A figure like figure \ref{kickspeed} can be obtained
after multiple tries.

A first approach could be limiting the kicks to several classes of kicks (10\%, 20\%, 30\% etc.), and estimate
for each class what is the average speed. Then, we could reverse this mapping to decide the target speed we want
to give to the ball. More sophisticated methods can be used here, like using least squares to fit a linear model
(like a polynomial function) over the obtained samples.

Another interesting point here is that we can also notice the noisy nature of our sampling, due to the fact that
we can't exactly reproduce the same initial conditions. This noise can also be estimated using a model of the
variance, which can later be useful to estimate not only where the ball will arrive, but a probability
density of where it will arrive.

\section{Competition}

Competition is one of the often mentioned driving forces of educational robotics, since it embeds the pedagogical
purposes into a project-oriented motivating schedule.

In this section, we describe how a competition would be designed with the RSK. The approach we propose allows
competitions to be performed with students from scratch on a fast schedule.

\subsection{Provided localization}

The architecture presented on figure \ref{architecture} is competition-ready. Setting up and operating
the \textit{Game Controller} can be the responsibility of the organizers. The competitors can use
their computers and join the same network as the \textit{Game Controller}.

Each team will then be given a key allowing them to control the robots for theirs color and preventing them from
controlling opponent's robots. All the localization information (robots and ball) will be published to all the
teams. The knowledge of the game state is then the same for everyone.

\subsection{Standard robots}

As explained before, most robotics competitions focus on the robot's hardware development by teams. We believe that this
is a very important aspect of robotics since it allows the platforms themselves to improve year after year. However,
it can hardly be addressed in general tracks in most high school. This is unfortunate since programming
the robots behaviors and motion are immediate applications of materials studied (physics, geometry or
computer sciences).

Having standard robots clearly changes the focus of a competition from design, low-level and embedded systems to
pure software concerns. This approach was for {\em e.g} adopted by the RoboCup major Standard Platform League (SPL)
\cite{chown2014standard} where humanoid soccer teams compete using only Aldebaran's Nao robots.

The FIRST LEGO league\footnote{http://firstlegoleague.org/} is a robotics competition with this ideology, where
students are asked to tackle challenges that change every year using standard LEGO parts.

RSK robots are open-source, meaning that it is possible for anybody to rebuild them from the documentation. However,
in a competition setup, we can imagine that the robots used by the players would not be their own, but provided
by the competition organizers. Thus, it would not be necessary to check the robots compliance with a rulebook.

We even suggest that robots are exchanged at half-game, allowing for perfect fairness. This only requires swapping
the robot's fiducial markers.

\subsection{Implementing rules in the Game Controller}

The fact that every piece of data flows through the \textit{Game Controller} allows a lot of technical simplifications.
Particularly, robots can be preempted: their motion is stopped and their control is prevented, or they can even be
moved to a wanted position. For instance, the game engagement placement can be managed before the teams actually
have the control of their robots.

One of the most important rule to implement is avoiding the robots to block the ball by staying very close to it for
a long time, and encourage the concept of placement and kicking. For that purpose, we propose that the time spent
close to the ball (in a small fixed radius around it) is limited by the rules. This is a typical example of something
that can be achieved inside the \textit{Game Controller} itself, enforcing this rule by penalizing automatically
robots violating it.

Another example is detecting goals. This can be done by the \textit{Game Controller} by default since it is only
a line intersection calculation. This is convenient since it allows playing without goal posts, which can
create other mechanical issues because of collisions with the robots.

It is also possible for the \textit{Game Controller} to log all the data sent and received during a game to view
replays after the games, which can help the teams to investigate their code between games.

\section{Conclusion and future work}

The implementation of a block coding interface  (for example Scratch) will be put in place in order to open the
use of the kit to elementary classes. This type of coding could facilitate the apprehension of the robots by students new to programming. In addition, an improvement path is also being considered to completely deport the server-side programming interface like Jupyter.


In order to apply all the educational elements presented in \ref{pedasequ}, we also decided to launch an experiment with 10 schools
in conjunction with the local education authority of the Nouvelle-Aquitaine region. 
The purpose of the study is double : on the one hand, to determine the attraction of these robots on students and thus the 
impact on their learning of computer science or engineering sciences (STEM). On the other hand, to determine the positive or negative 
impact of a competitive context in a school environment by involving students in a soccer competition with these robots.

In order to carry out all this, kits as presented in Figure \ref{kit} were provided to teachers of each high school. 
The classes involved have a representative panel of students from different levels
\footnote{from 9th to 12th Grade (American education system)} and sections\footnote{Engineering sciences, IT, general education.}.
This experiment has now started and is currently in progress as of this writing.

Moreover, in addition to this project, a virtual simulator of the robot soccer kit is under development in order to study the impact 
of the real system against a simulated environment in a learning context. Indeed, this second study will be carried out jointly with 
the local education authority which is already involved in the experiment in progress.

\bibliographystyle{IEEEtran}
\bibliography{biblio}

\begin{thebibliography}{10}
\providecommand{\url}[1]{#1}
\csname url@samestyle\endcsname
\providecommand{\newblock}{\relax}
\providecommand{\bibinfo}[2]{#2}
\providecommand{\BIBentrySTDinterwordspacing}{\spaceskip=0pt\relax}
\providecommand{\BIBentryALTinterwordstretchfactor}{4}
\providecommand{\BIBentryALTinterwordspacing}{\spaceskip=\fontdimen2\font plus
\BIBentryALTinterwordstretchfactor\fontdimen3\font minus
  \fontdimen4\font\relax}
\providecommand{\BIBforeignlanguage}[2]{{%
\expandafter\ifx\csname l@#1\endcsname\relax
\typeout{** WARNING: IEEEtran.bst: No hyphenation pattern has been}%
\typeout{** loaded for the language `#1'. Using the pattern for}%
\typeout{** the default language instead.}%
\else
\language=\csname l@#1\endcsname
\fi
#2}}
\providecommand{\BIBdecl}{\relax}
\BIBdecl

\bibitem{evripidou2020educational}
S.~Evripidou, K.~Georgiou, L.~Doitsidis, A.~A. Amanatiadis, Z.~Zinonos, and
  S.~A. Chatzichristofis, ``Educational robotics: Platforms, competitions and
  expected learning outcomes,'' \emph{IEEE access}, vol.~8, pp.
  219\,534--219\,562, 2020.

\bibitem{resnick2009scratch}
M.~Resnick, J.~Maloney, A.~Monroy-Hern{\'a}ndez, N.~Rusk, E.~Eastmond,
  K.~Brennan, A.~Millner, E.~Rosenbaum, J.~Silver, B.~Silverman \emph{et~al.},
  ``Scratch: programming for all,'' \emph{Communications of the ACM}, vol.~52,
  no.~11, pp. 60--67, 2009.

\bibitem{kitano1997robocup}
H.~Kitano, M.~Asada, Y.~Kuniyoshi, I.~Noda, E.~Osawa, and H.~Matsubara,
  ``Robocup: A challenge problem for ai,'' \emph{AI magazine}, vol.~18, no.~1,
  pp. 73--73, 1997.

\bibitem{sklar2002robocupjunior}
E.~Sklar, A.~Eguchi, and J.~Johnson, ``Robocupjunior: learning with educational
  robotics,'' in \emph{Robot Soccer World Cup}.\hskip 1em plus 0.5em minus
  0.4em\relax Springer, 2002, pp. 238--253.

\bibitem{weitzenfeld2014robocup}
A.~Weitzenfeld, J.~Biswas, M.~Akar, and K.~Sukvichai, ``Robocup small-size
  league: Past, present and future,'' in \emph{Robot Soccer World Cup}.\hskip
  1em plus 0.5em minus 0.4em\relax Springer, 2014, pp. 611--623.

\bibitem{lynch2017modern}
K.~M. Lynch and F.~C. Park, \emph{Modern robotics}.\hskip 1em plus 0.5em minus
  0.4em\relax Cambridge University Press, 2017.

\bibitem{romero2018speeded}
F.~J. Romero-Ramirez, R.~Mu{\~n}oz-Salinas, and R.~Medina-Carnicer, ``Speeded
  up detection of squared fiducial markers,'' \emph{Image and vision
  Computing}, vol.~76, pp. 38--47, 2018.

\bibitem{opencv_library}
G.~Bradski, ``{The OpenCV Library},'' \emph{Dr. Dobb's Journal of Software
  Tools}, 2000.

\bibitem{dubrofsky2009homography}
E.~Dubrofsky, ``Homography estimation,'' \emph{Diplomov{\'a} pr{\'a}ce.
  Vancouver: Univerzita Britsk{\'e} Kolumbie}, vol.~5, 2009.

\bibitem{chown2014standard}
E.~Chown and M.~G. Lagoudakis, ``The standard platform league,'' in \emph{Robot
  Soccer World Cup}.\hskip 1em plus 0.5em minus 0.4em\relax Springer, 2014, pp.
  636--648.

\end{thebibliography}

\end{document}